\documentclass[runningheads]{llncs}

\usepackage[T1]{fontenc}
\usepackage{graphicx}

\usepackage[table]{xcolor}
\usepackage[most]{tcolorbox}
\usepackage{booktabs}
\usepackage{tabularx}
\usepackage{array}
\usepackage{siunitx}
\usepackage{subcaption}
\usepackage{tikz}

\usepackage{enumitem}
\usepackage{pgfplots}
\pgfplotsset{compat=1.18}
\usepgfplotslibrary{polar}
\usepackage{booktabs}
\usepackage{multirow}



\definecolor{gold}{rgb}{1.0, 0.84, 0}
\definecolor{silver}{rgb}{0.8, 0.8, 0.8}
\definecolor{bronze}{rgb}{0.8, 0.5, 0.2}

\tcbset{
tagbase/.style={
on line,
fontupper=\bfseries\footnotesize,
boxsep=2pt,
left=5pt,
right=5pt,
top=2pt,
bottom=2pt,
arc=4pt,
boxrule=0.6pt
},
modelchipbase/.style={
on line,
fontupper=\ttfamily\footnotesize,
boxsep=1pt,
left=3pt,
right=3pt,
top=1pt,
bottom=1pt,
arc=3pt,
boxrule=0.4pt
}
}

\newtcbox{\tagModelType}{tagbase, colback=green!15!white, colframe=green!60!black}
\newtcbox{\tagLexical}{tagbase, colback=gray!15!white, colframe=gray!70!black}
\newtcbox{\tagTechniques}{tagbase, colback=cyan!10!white, colframe=cyan!70!black}
\newtcbox{\tagLoss}{tagbase, colback=orange!15!white, colframe=orange!80!black}
\newtcbox{\tagLLM}{tagbase, colback=purple!15!white, colframe=purple!70!black}
\newtcbox{\tagExternalData}{tagbase, colback=brown!10!white, colframe=brown!60!black}
\newtcbox{\tagBias}{tagbase, colback=blue!15!white, colframe=blue!50!black}

\newtcbox{\tagModel}{modelchipbase, colback=black!4!white, colframe=black!45!white}

\newtcbox{\tagLabelModelType}{on line, colback=green!15!white, colframe=green!60!black,
fontupper=\small, boxsep=1pt, left=2pt, right=2pt, top=1pt, bottom=1pt, arc=4pt, boxrule=0.5pt}

\newtcbox{\tagLabelTechniques}{on line, colback=cyan!10!white, colframe=cyan!70!black,
fontupper=\small, boxsep=1pt, left=2pt, right=2pt, top=1pt, bottom=1pt, arc=4pt, boxrule=0.5pt}

\newtcbox{\tagLabelLoss}{on line, colback=orange!15!white, colframe=orange!60!black,
fontupper=\small, boxsep=1pt, left=2pt, right=2pt, top=1pt, bottom=1pt, arc=4pt, boxrule=0.5pt}

\newtcbox{\tagLabelLLM}{on line, colback=purple!15!white, colframe=purple!70!black,
fontupper=\small, boxsep=1pt, left=2pt, right=2pt, top=1pt, bottom=1pt, arc=4pt, boxrule=0.5pt}

\newtcbox{\tagLabelBias}{on line, colback=blue!15!white, colframe=blue!50!black,
fontupper=\small, boxsep=1pt, left=2pt, right=2pt, top=1pt, bottom=1pt, arc=4pt, boxrule=0.5pt}

\newtcbox{\tagLabelExternalData}{on line, colback=brown!10!white, colframe=brown!60!black,
fontupper=\small, boxsep=1pt, left=2pt, right=2pt, top=1pt, bottom=1pt, arc=4pt, boxrule=0.5pt}

\newtcbox{\tagLabelModel}{on line, colback=black!4!white, colframe=black!45!white,
fontupper=\ttfamily\scriptsize, boxsep=1pt, left=2pt, right=2pt, top=1pt, bottom=1pt, arc=3pt, boxrule=0.4pt}

\begin{document}
\title{Overview of the TalentCLEF 2026: Skill and Job Title Intelligence for Human Capital Management}
\titlerunning{TalentCLEF at CLEF2026}
\author{
Luis Gasco\inst{1}\orcidID{0000-0002-4976-9879}\thanks{Equal contribution.} \and
Hermenegildo Fabregat\inst{1}* \and
Laura García-Sardiña\inst{1}* \and
Paula Estrella\inst{1}* \and
Warre Veys\inst{2}* \and
Casimiro Pío Carrino\inst{1} \and
Matthias De Lange\inst{2} \and
Daniel Deniz Cerpa\inst{1} \and
Álvaro Rodrigo\inst{3} \and
Jens-Joris Decorte\inst{2} \and
Rabih Zbib\inst{1}
}
\authorrunning{L. Gasco et al.}
%
\institute{
Avature Machine Learning, Spain \\
\email{machinelearning@avature.net}
\and
TechWolf, Belgium
\and
NLP \& IR Group at UNED, Madrid, Spain
}

%
\maketitle              
\begin{abstract}
This paper presents an overview of the second edition of the TalentCLEF challenge, organized as a Lab at the Conference and Labs of the Evaluation Forum (CLEF) 2026. TalentCLEF is an initiative aimed at advancing Natural Language Processing research in Human Capital Management. The second edition of the challenge consisted of two tasks: Task A, contextualized job-person matching, focuses on identifying and ranking the most suitable candidates represented by their resumes for a given job vacancy in English and Spanish. Task B, job-skill matching with skill type classification, addresses retrieving the most relevant skills for a given job title in English and distinguishing between core and contextual skills. TalentCLEF attracted 113 registered teams and received more than 400 submissions in the two tasks, reflecting the growing interest of the research community in shared evaluation benchmarks for Human Capital Management. This paper describes the motivation and organization of the challenge, summarizes the datasets and evaluation settings, and reports the main results obtained by the participating teams.

\end{abstract}
%
%
%

\section{Introduction}

The transformation of the labor market is changing the way organizations describe jobs, identify talent, and support career development~\cite{linkedin2025skillsisgnal}. In this context, Human Capital Management (HCM) increasingly requires Natural Language Processing (NLP) systems capable of processing and connecting information about jobs, people, and skills. Such systems are relevant not only for recruitment and talent acquisition, but also for broader workforce development scenarios, including career guidance, skill gap analysis, internal mobility, upskilling, and reskilling.

Language technologies are particularly well suited to address these challenges since much of the relevant information in this domain is expressed in text. Job advertisements, job titles, professional profiles, curricula, learning resources, and skill or occupation taxonomies contain valuable information about the relationships between occupations, skills, workers, and learning opportunities. By extracting, normalizing, matching, and linking this information, NLP methods can support more structured representations of the labor market and enable downstream applications for job and skill intelligence.

In recent years, the application of NLP to Human Capital Management has received growing research attention, as reflected in dedicated venues such as \textit{NLP4HR}~\cite{hruschka2024proceedings} and \textit{RecSys in HR}~\cite{bogers2024fourth}. These initiatives have helped consolidate the area by bringing together research on traditional NLP tasks applied in the Human Resources (HR) area, such as skill extraction from job postings~\cite{zhang2022skill,zhang2022skillspan,nguyen2024rethinking,vermeer2022using}, skill normalization to taxonomies~\cite{de2026skillens,retyk2024melo}, matching~\cite{decorte2024skillmatch,lavi2021consultantbert,deniz2024combined}, job recommendation~\cite{giabelli2021skills2job,decorte2021jobbert,anand2022required,fabregat2024inductive}, but also novel ones such as career path modeling~\cite{decorte2023career,sathish2024significance}, LLM comprehension~\cite{carrino2026jobresqa,vijayalakshmi2024optimization}, or the analysis of fairness and bias in recruitment-related systems~\cite{arafan2022end,garcia2025measuring,pena2025addressing,sivakaminathan2026chatgpt}.

Despite recent advances in this area, research remains fragmented. Existing studies often rely on different datasets, languages, task definitions, annotation schemes, and evaluation protocols, making it difficult to compare systems or assess progress consistently. This fragmentation is particularly problematic given that such systems can influence real-world decision-making scenarios, including recruitment. Therefore, the development of public benchmarks remains an important need in the field, as they can help structure progress in similar ways to previous initiatives in other domains, such as biomedical NLP~\cite{nentidis2024bioasq}.

TalentCLEF addresses this fragmentation by providing a shared evaluation framework for NLP systems in Human Capital Management~\cite{gasco2025overview,gasco2025talentclef}. The initiative is organized around competitive evaluation campaigns grounded in realistic job and skill intelligence scenarios, with the goal of promoting the development of robust, multilingual, and reusable language technologies. At the same time, TalentCLEF provides the research community with high-quality datasets, common evaluation protocols, and public benchmarks that support reproducible research and future system comparison.

In this paper, we present an overview of the TalentCLEF 2026 Challenge. We describe the motivation of the challenge, introduce the proposed tasks, and summarize the main results obtained by the participating systems. The challenge attracted 113 registered teams and received more than 400 submissions across the two tasks. We also analyze the main methodological trends observed in the submitted systems, including hybrid retrieval, reranking approaches, generative AI components, and the use of structured knowledge sources such as skill and occupation graphs. Detailed descriptions of each task, including dataset construction, annotation procedures, and task-specific evaluation settings, are provided in the corresponding task overview articles~\cite{jobperson2026talentclef,jobskill2026talentclef}.

\section{Overview of the Tasks}

The second edition of the TalentCLEF Challenge~\cite{gasco2026talentclef} aims to promote the development and evaluation of systems for two highly relevant tasks in Human Capital Management: Task A, candidate search for a given job vacancy, and Task B, identification of professional skills relevant to specific job positions.

\subsection{Task A - Contextualized Job-Person Matching}
Candidate matching is one of the main challenges in Human Capital Management. When performed manually, this process typically relies on the individual reading of the resumes, the interpretation of the job descriptions, and the expertise and judgment of the recruiters. Although this approach allows human judgment and contextual knowledge to be incorporated into the decision-making process, it is increasingly difficult to scale in today's labor market, where a single job vacancy may receive hundreds of candidate profiles~\cite{khelkhal2025smart}. As a result, manually reviewing such large volumes of applications is often not feasible in practice.

In response to these limitations, automatic matching systems based on NLP and information extraction techniques have been developed in recent years. These systems usually focus on identifying and normalizing relevant entities in both job vacancies and candidate profiles, such as job titles, skills, competencies, education, and languages~\cite{zbib2022learning,garcia2023normalisation}. Once this information has been extracted, the systems compare the entities present in both types of document to estimate the degree of suitability between a candidate and a job vacancy. This represents a valid and widely used approach. However, the recent emergence of Large Language Models (LLMs) opens new possibilities for addressing the matching problem from a richer and more contextual perspective. In particular, LLM-based approaches can help incorporate additional dimensions into the matching process without requiring specific fine-tuning, such as seniority level, evidence of expertise in key skills, demonstrated experience in real-world contexts, and the inference of implicit or related skills, among others~\cite{vaishampayan2025human,ghosh2023jobrecogpt,lo2025ai}. This creates new opportunities for the development of more flexible and context-aware candidate matching systems.

In the previous edition of TalentCLEF Task A, the problem focused exclusively on job title matching. However, as discussed above, candidate matching involves many other types of information, which can be extracted from documents containing much richer descriptions of both candidates and job vacancies. For this reason, Task A in this year's edition has been framed as a broader problem of contextualized job-person matching, where the aim is to develop systems capable of identifying and ranking the most suitable candidates for a given job vacancy. To support this task, we provide a manually-annotated synthetic corpus composed of job description and candidate profiles. Participants are free to process this corpus using different methodological approaches, including information extraction, prompt engineering, information retrieval, or other NLP-based techniques, to generate a ranked list of candidates for each job vacancy. 

\begin{figure}[h!]
    \centering
    \includegraphics[width=0.75\textwidth]{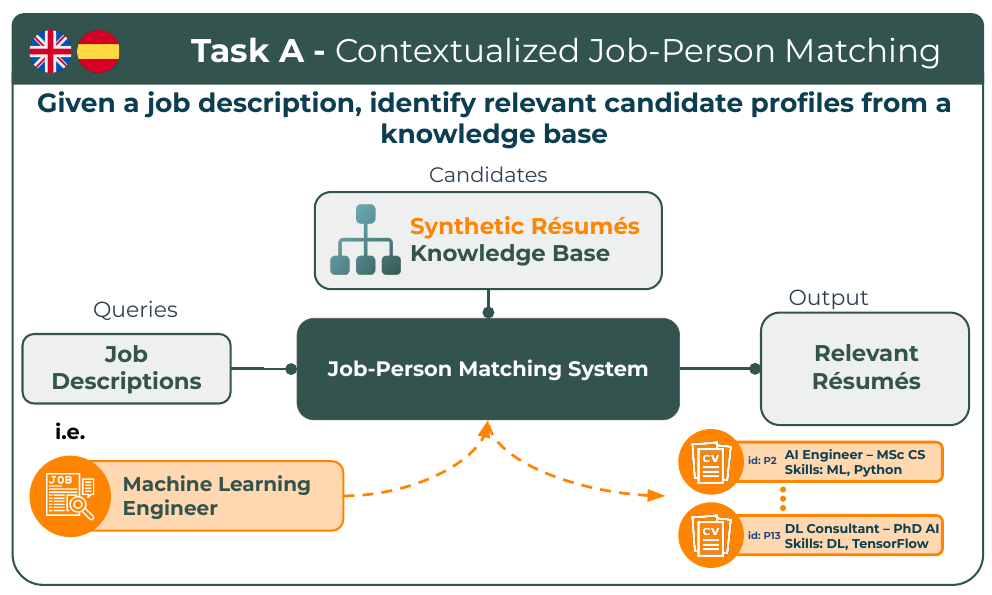}
    \caption{Overview of Task A: Contextualized Job–Person Matching}
    \label{fig:my_image}
\end{figure}

\subsubsection{Data}
In this task, we provide a multilingual dataset for contextualized person-job matching, covering both English and Spanish. The task corpus is divided into development and test set. Training data were not provided for the task; however, participating teams were allowed to use any external resources or additional information they considered relevant.

Both partitions were synthetically generated, including job records and resumes that describe vacancy requirements and candidate profiles. Rather than relying on uncontrolled generation, the process was guided by statistical evidence on job-skill co-occurrence patterns extracted from real-world resumes and job descriptions, obtained from an internal database\footnote{Although we used real data for the data generation, a manual review on the statistical evidence was done for avoiding data leakage.}. The selection of profiles and vacancies was designed to cover a wide range of industries, occupations, professional backgrounds, gender, and ethnicity. The resulting candidate-job pairs were subsequently reviewed and manually annotated by expert annotators for the matching task.

Table~\ref{tab:corpus-taska} summarizes the main statistics of the corpus, including the number of job descriptions (queries), resumes (corpus) per language. Data for this task were made available to participating teams through Zenodo~\footnote{Task A corpus: https://doi.org/10.5281/zenodo.17625261}. Further details on the dataset generation and annotation process are provided in the extended overview of the task.

\begin{table}[ht!]
\centering
\caption{Statistics of Task A's development and test sets by language.}
\label{tab:corpus-taska}
\begin{tabular}{llrr}
\toprule
\textbf{Split} & \textbf{Language} & \textbf{Queries} & \textbf{Corpus} \\
\midrule
\multirow{2}{*}{Development}
    & English & 10 & 472 \\
    & Spanish & 10 & 472 \\
\midrule
\multirow{2}{*}{Test}
    & English & 40 & 476 \\
    & Spanish & 40 & 476 \\
\bottomrule
\end{tabular}
\end{table}

\subsubsection{Evaluation}
The task was evaluated through a competition hosted on Codabench~\footnote{Task A Codabench: https://www.codabench.org/competitions/14226/}, which provided the participating teams with a common environment to submit their predictions and access the official leaderboards. In addition, the use of this platform allows the task to remain available as an open benchmark for continuous evaluation after the end of the challenge, supporting reproducibility and the comparison of future systems.

In this edition, we consider three evaluation settings. The first is multilingual, in which both the job vacancy and the candidate resumes are written in the same language. The second is cross-lingual, with the job vacancy written in English and the resumes in Spanish. This setting is relevant in multilingual contexts, where a company may publish a vacancy in one language while candidates describe their work experience in another. The third evaluation focuses on bias. Since job matching systems can have a direct impact on people’s employment opportunities, it is essential to analyze not only their overall performance, but also their behavior between demographic groups, such as gender or ethnicity. In this case, we assess whether the systems produce consistent and fair rankings regardless of the candidate's gender.

For monolingual and cross-lingual scenarios, system performance is measured using Mean Average Precision (MAP) over the ranked list of candidates. For the bias scenario, we use the Rank-Biased Overlap (RBO) to evaluate gender bias.

\subsection{Task B - Job-Skill Matching with Skill Type Classification}

Skills have become a central component of HCM. In recent years, the emergence of artificial intelligence and other technological transformations has accelerated changes in the labor market: new job roles are appearing at an unprecedented pace, existing occupations are being rapidly redefined, and the skill requirements associated with many positions are continuously evolving. As a result, organizations increasingly need support systems not only to define new professional roles and the technical skills they require, but also to update the knowledge and capabilities of their workforce so that employees can progressively adapt to technological change.

This shift has reinforced the importance of skill-based approaches in recruitment, workforce planning, and talent development. In recruitment, such systems can help identify candidates whose capabilities match the requirements of a role, even when their previous job titles or career paths are not directly related to the vacancy. In workforce management, they can support the identification of skill gaps and the recommendation of learning pathways that help workers adapt to new occupational demands.

Last year, Task B focused on retrieving the skills most relevant to a given job title. This year, the task expands that setting by requiring systems not only to identify relevant skills, but also to consider whether each skill is core or contextual for the target job. Core skills are those required to perform a job regardless of the work context or employer and are therefore essential to the position. Contextual skills, on the contrary, depend on factors such as industry, organization, or a specific work environment and can be considered complementary or optional depending on the context.

The objective of Task B this year is to develop systems capable of understanding both the relevance and the role of professional skills in relation to job titles. Given a database of professional skills and a specific job, participating systems are required to identify the most relevant skills and classify them according to their importance for the position. 

\begin{figure}[h!]
    \centering
    \includegraphics[width=0.75\textwidth]{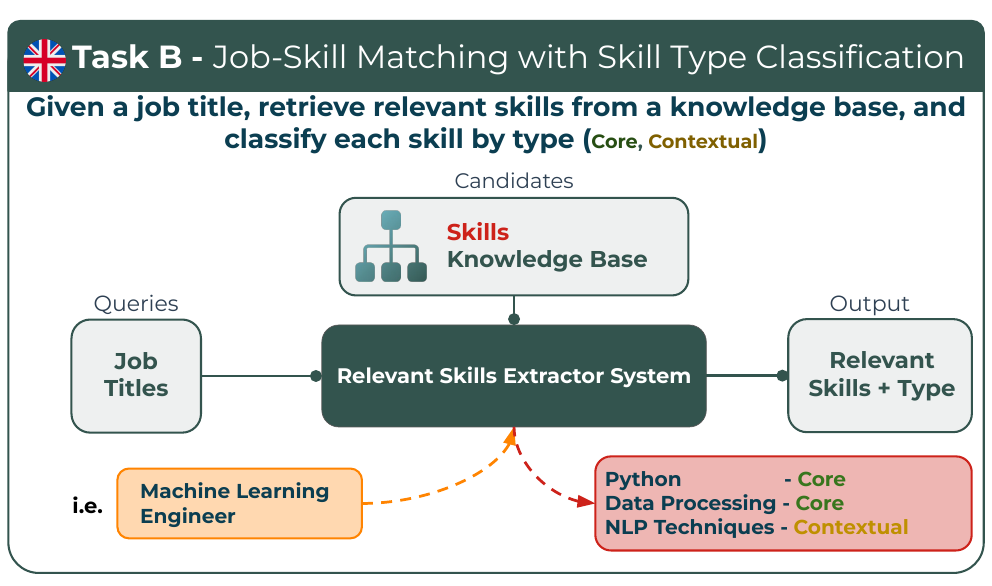}
    \caption{Overview of Task B: Job-Skill Matching with Skill Type Classification}
    \label{fig:my_image}
\end{figure}

\vspace{-1.5em}

\subsubsection{Data}

In Task B, we provide a monolingual corpus in English divided into three splits: a training set, a development set, and a test set. The training data consist of a list of the most representative skills for each ESCO occupation. A filtering process was applied to limit the number of skills associated with each job title and avoid outlier cases with an unusually large number of skills. These data also specify whether each skill is essential or optional for the corresponding occupation.

For the development and test sets, we refined and expanded the dataset used in last year's edition of the task. The queries corresponding to job titles are the same as in the previous edition, but the corpus elements were expanded and revised to provide a richer and more reliable evaluation setting. This process combined semantic techniques, human validation, and LLM-as-a-judge procedures to improve coverage and ensure the quality of relevance annotations.

Table~\ref{tab:data-stats-taskb} summarizes the main statistics of the corpus, including the number of queries, the corpus elements, and the average number of relevant items per query. Data were made available through Zenodo~\footnote{Task B corpus: https://doi.org/10.5281/zenodo.17625261}. Further details on the generation and validation of the evaluation dataset are provided in the extended overview of the task.

\vspace{-1em}

\begin{table}[ht!]
\centering
\caption{Statistics of Task B's development and test sets.}
\label{tab:data-stats-taskb}
\begin{tabular}{lrr}
\hline
\textbf{Statistic} & \textbf{Development} & \textbf{Test} \\
\hline
Number of queries                 & 304   & 125   \\
Number of corpus elements         & 9,052 & 5,358 \\
Average relevant items per query  & 185.6 & 184.2 \\
\hline
\end{tabular}
\label{tab:dataset-statistics}
\end{table}

\vspace{-2.5em}

\subsubsection{Evaluation}
The evaluation of Task B was conducted using a Codabench competition~\footnote{Codabench Task B: https://www.codabench.org/competitions/14489/}. The main evaluation metric was Normalized Discounted Cumulative Gain (NDCG), which assesses ranking quality by considering both the relevance of the retrieved items and their position in the ranked list.

Although the task includes a skill-type classification component, this distinction is incorporated into the ranking evaluation through two relevance settings. In the binary relevance scenario, all relevant skills are treated equally: both core and contextual skills receive the same relevance value. This setting evaluates whether systems can retrieve skills that are relevant to a given job title, regardless of their type.

In the graded relevance scenario, the evaluation also accounts for the type of each relevant skill. Core skills are assigned a higher relevance value of 2, while contextual skills are assigned a relevance value of 1. This setting rewards systems that rank core skills higher, reflecting their greater importance in performing a specific occupation, while still giving credit for retrieving contextual skills that may be relevant depending on the work environment.

\section{Participants}

\subsection{Task A}

Task A attracted 91 registered participants this year. During the evaluation phase, 21 teams submitted at least one run, producing a total of 384 submissions across the development and test sets, with 12 teams sending working notes to be included in the official benchmark. Table~\ref{tab:participant_task_a_2026} summarizes the main systems and approaches used in this task by the participating teams. \tagLabelModelType{Representation type} describes the type of text representation used in their submissions, while \tagLabelTechniques{Methodology} captures the main methodological strategies adopted to address the task, such as retrieval or reranking. \tagLabelLoss{System details} reports additional information about the submitted systems, including learning objectives, fine-tuning strategies, rank fusion methods, graph-based components, or other relevant architectural choices. \tagLabelLLM{LLM-related} identifies the role of LLMs in the submitted systems. \tagLabelBias{Bias mitigation} indicates whether specific mechanisms were introduced to address gender bias, and \tagLabelExternalData{External data} reports the use of additional resources beyond the data provided by the organizers.

\begin{table}[htbp]
\centering
\caption{Overview of participating team approaches for Task A. }
\label{tab:participant_task_a_2026}
\resizebox{\textwidth}{!}{%
\begin{tabular}{@{}p{2.7cm}lp{8.5cm}p{7.2cm}@{}}
\toprule
\textbf{Team} & \textbf{Ref} & \textbf{Methods} & \textbf{Models} \\
\midrule

qwerity &
\cite{palomino2026qwerity} &
\tagModelType{Encoder}\tagModelType{Lexical}\tagTechniques{Retrieval}\tagTechniques{Reranking}
\tagLoss{Late-Interaction reranking} &
\tagModel{JobBERT-v3}
\tagModel{SPLADE}
\tagModel{jina-colbert-v2} \\

\midrule

TalentCLEF Task 1 -- Methodology 5 -- Group 1 &
\cite{hamza2026talentcleftask1methodology5group1} &
\tagModelType{Encoder}\tagTechniques{Retrieval}\tagTechniques{Reranking} 
\tagLoss{Cross-Encoder reranking} &
\tagModel{cross-encoder/mmarco-mMiniLMv2-L12-H384-v1}
\tagModel{paraphrase-multilingual-MiniLM-L12-v2}
\tagModel{paraphrase-multilingual-mpnet-base-v2} \\

\midrule

CYUT &
\cite{poerwanto2026cyut} &
\tagModelType{Encoder}\tagModelType{Decoder}\tagModelType{Lexical}\tagTechniques{Retrieval}
\tagTechniques{Reranking}\tagLoss{Weighted RRF}\tagExternalData{External data}
\tagLLM{Hypothetical Document Embeddings Query Expansion}
\tagLLM{Prompt engineering}\tagLLM{Listwise reranking}
 &
\tagModel{bge-m3}
\tagModel{Gemma-3-27B}
\tagModel{BM25} \\

\midrule

classum &
\cite{kim2026classum} &
\tagModelType{Encoder}\tagModelType{Decoder}\tagTechniques{Retrieval}\tagTechniques{Reranking}
\tagTechniques{Rank fusion}\tagLLM{Prompt engineering}
\tagLLM{LLM entity extraction}\tagLLM{Listwise reranking} &
\tagModel{JobBERT-v3}
\tagModel{KaLM-12B}
\tagModel{Octen-8B}
\tagModel{Nemotron-8B}
\tagModel{ZEmbed1-4B}
\tagModel{Harrier-27B}
\tagModel{gemini-embedding-001}
\tagModel{GPT-5.5}
\tagModel{Claude Opus 4.6}
\tagModel{Gemini 3 Flash}
\tagModel{Cohere-rerank-4}
\tagModel{zerank-2} \\

\midrule

DevoMatcher &
\cite{riabi2026devomatcher} &
\tagModelType{Encoder}\tagModelType{Decoder}\tagModelType{Lexical}\tagTechniques{Retrieval}
\tagTechniques{Reranking}\tagLoss{ESCO graph calibration score}
\tagLoss{Weighted RRF}\tagLLM{Listwise reranking}
\tagBias{Demographic Masking} &
\tagModel{BM25}
\tagModel{BGE-M3}
\tagModel{SkillMatch MPNet}
\tagModel{OpenAI GPT-5.2} \\

\midrule

Skillberg.app &
\cite{vielvoye2026skillberg} &
\tagModelType{Encoder}\tagModelType{Lexical}\tagTechniques{Retrieval}\tagTechniques{Reranking}
\tagLoss{Hybrid retrieval}\tagLoss{Curriculum Learning}
\tagLoss{GISTEmbedLoss}\tagLoss{MatryoshkaLoss}
\tagLoss{Proprietary job-skill knoweledge graph}
\tagLoss{Cross-Encoder reranking}
\tagLLM{Prompt engineering} \tagLLM{Data Augmentation} &
\tagModel{multilingual-e5-large-instruct }
\tagModel{BM25}
\tagModel{Mistral}
\tagModel{Jina cross-encoder v2} \\

\midrule

HR\_NLP &
\cite{dumitru2026hrnlp} &
\tagModelType{Encoder}\tagTechniques{Retrieval}\tagLoss{Triplet margin loss}
\tagLoss{teacher-student Knowledge distillation}
\tagLoss{Student Contrastive Loss}
\tagExternalData{External data} &
\tagModel{JobBERT}
\tagModel{JobBERT-v3}
\tagModel{RoBERTa} \\

\midrule

UBCS &
\cite{thuma2026unionfusion} &
\tagModelType{Encoder}\tagModelType{Encoder--Decoder}\tagModelType{Lexical}
\tagTechniques{Retrieval}\tagLoss{Late fusion}\tagLoss{Doc2Query expansion} &
\tagModel{paraphrase-multilingual-MiniLM-L12-v2 }
\tagModel{doc2query-t5-base-msmarco}
\tagModel{BM25} \\

\midrule

VerbaNex &
\cite{moreno2026verbanex} &
\tagModelType{Encoder}\tagModelType{Decoder}\tagModelType{Lexical}\tagTechniques{Retrieval}
\tagTechniques{Reranking}\tagLoss{Cross-Encoder reranking}
\tagLoss{Score fusion}\tagLLM{Prompt engineering}
\tagLLM{Query expansion}&
\tagModel{bge-m3}
\tagModel{BM25}
\tagModel{mMiniLMv2 cross-encoder} \\

\midrule

ui-nlp &
\cite{luthfiyyah2026uinlp} &
\tagModelType{Encoder}\tagModelType{Lexical}\tagTechniques{Retrieval}\tagTechniques{Reranking}
\tagLoss{Rank Fusion} &
\tagModel{multilingual-e5-large-instruct}
\tagModel{bge-large-en-v1.5}
\tagModel{BM25}
\tagModel{ms-marco-MiniLM-L12-v2 cross-encoder} \\

\midrule

bipboopbipboop &
\cite{hemamou2026retrievereranksurvive} &
\tagModelType{Encoder}\tagModelType{Decoder}\tagTechniques{Retrieval}\tagTechniques{Reranking}
\tagLoss{Geometric-mean fusion}\tagLoss{Cosine similarity Loss}
\tagLoss{LoRA}\tagLLM{Prompt engineering}\tagLLM{LLM-as-a-judge}
\tagLLM{Pointwise reranking}
\tagLLM{Tournament listwise  reranker}
&
\tagModel{Qwen3-Reranker-8B}
\tagModel{Qwen3-235B}
\tagModel{BGE-LoRA}
\tagModel{MiniLM} \\

\midrule

QTPride &
\cite{tran2026qtpride} &
\tagModelType{Encoder}\tagTechniques{Retrieval}\tagLoss{RFF}
\tagLoss{Relational Graph Convolutional Network}
\tagLoss{KG-enhanced retrieval}
\tagLLM{LLM entity extraction} &
\tagModel{multilingual-e5-large-instruct}\\

\bottomrule
\end{tabular}
}
\end{table}

%

The solutions proposed this year for Task A are predominantly based on multi-stage information retrieval architectures. Participants consistently frame the problem as a retrieval task: they first generate an initial set of candidates using diverse textual representation methods and, in many cases, apply a subsequent reranking stage to refine the final ranking


In the retrieval stage, models that generate dense semantic representations constitute the core component of the systems, but they are often complemented with lexical methods such as BM25, which appears in several submissions as an exact-matching mechanism. This might be particularly useful in the HCM domain, where normalized skill names and labor-market expressions can play an important role in retrieving a good set of candidates. In addition, models from the JobBERT family~\cite{decorte2021jobbert} are used by teams such as \textit{qwerity}, \textit{classum}, and \textit{HR\_NLP}, trying to take advantage of models adapted to the labor-market domain compared with more general-purpose embedding models.


A particularly frequent pattern is the incorporation of reranking stages in the retrieved candidates. Nine out of the 12 teams explicitly include some sort of reranking mechanism. \textit{qwerity}, for example, explores different strategies by combining lexical models such as SPLADE~\cite{lassance2024splade} with late-interaction reranking based on ColBERT architectures~\cite{khattab2020colbert}. Other teams, such as \textit{VerbaNex}, \textit{ui-nlp}, and \textit{Skillberg.app}, use pretrained cross-encoders to improve the precision of the final ranking.

Because participants often combine different representations to produce the final ranking, rank fusion methods are a central method in many systems. Techniques such as reciprocal rank fusion, late fusion, score fusion, geometric-mean fusion, and other rank fusion variants are used to aggregate evidence from different rankings, enabling systems to combine complementary strengths: lexical models capture exact terminology, dense encoders capture semantic similarity, cross-encoders improve relevance estimation, and LLM-based rerankers, which are sometimes applied before and sometimes after rank fusion, provide more fine-grained relevance estimates. The widespread use of these methods highlights their importance for the task and suggests that integrating multiple sources of evidence often yields better results than relying on a single model.


One of the most relevant differences with respect to the first edition of the task is the more extensive use of LLMs. While the previous edition provided more limited contextual information, this year several teams exploited generative models and prompting techniques to evaluate or reorder candidates. \textit{CYUT}, \textit{DevoMatcher}, \textit{classum}, and \textit{bipboopbipboop} incorporated LLM-based reranking strategies. In particular, some systems used listwise reranking, where the model receives several candidates simultaneously and produces a comparative ordering. This approach contrasts with the pointwise reranking strategy used by \textit{bipboopbipboop}, where each candidate is evaluated independently rather than in comparison with the other candidates. This team also proposed an additional variant based on tournament-style listwise reranking, in which candidates are compared through successive rounds or pairwise competitions~\cite{chen2025tourrank}.

Beyond reranking, LLMs are also used for other tasks within the pipeline, mainly as a data augmentation technique. \textit{VerbaNex} uses them for query expansion, \textit{Skillberg.app} applies data augmentation, and \textit{CYUT} incorporates Hypothetical Document Embeddings for query expansion (HyDE)~\cite{li2025query}. These techniques aim to enrich the original input with additional information to improve the system’s ability to retrieve relevant candidates. In addition, \textit{bipboopbipboop} uses an LLM-as-a-judge approach to assess candidate adequacy during the retrieval stage, making it the only team to incorporate LLM knowledge directly into the retrieval task.

Two distinctive cases are \textit{QTPride} and \textit{classum}, which use LLM-based entity extraction as a step prior to retrieval. In addition, \textit{QTPride} combines the extracted entities with the information from the ESCO skill graph, which is fine-tuned using a Relational Graph Convolutional Network to learn structure-aware embeddings~\cite{schlichtkrull2018modeling}. This allows relational information between entities to be incorporated into the retrieval process. Other teams also exploit graph-based information: \textit{Skillberg.app} uses a proprietary knowledge graph of job-skill relations, while \textit{DevoMatcher} incorporates a calibration signal based on the ESCO graph to adjust the final ranking~\cite{le2014esco}. 


From a training perspective, not all teams perform task-specific fine-tuning of embedding models for the retrieval stage. However, when fine-tuning is applied, the strategies are varied. \textit{bipboopbipboop} uses LoRA techniques and cosine similarity loss to fine-tune their models~\cite{hu2022lora}. \textit{Skillberg.app} proposes a curriculum learning methodology, using GISTEmbedLoss as the loss function and MatryoshkaLoss to reduce the effective size of the vectors~\cite{wang2021survey,solatorio2024gistembed,kusupati2022matryoshka}. \textit{HR\_NLP}, in turn, uses a teacher-student knowledge distillation strategy to adapt their retrieval model. Finally, \textit{UBCS} employs encoder-decoder models to perform query expansion over development and test sets, reinforcing the idea that text generation can be used not only for ranking but also to enrich query representations.

\subsection{Task B}
Task B attracted 101 registered participants this year. During the evaluation phase, 15 teams submitted at least one run, producing a total of 43 submissions across the development and test sets. Table~\ref{tab:participant_task_b_2026} summarizes the main systems and approaches used by the participating teams in this task.


\begin{table}[ht!]
\centering
\caption{Overview of participating team approaches for Task B. }
\label{tab:participant_task_b_2026}
\resizebox{\textwidth}{!}{%
\begin{tabular}{@{}llp{8.5cm}p{7.2cm}@{}}
\toprule
\textbf{Team} & \textbf{Ref} & \textbf{Methods} & \textbf{Models} \\
\midrule

NightSun &
\cite{liu2026nightsun} &
\tagModelType{Encoder}\tagModelType{Decoder}\tagTechniques{Retrieval}\tagTechniques{Reranking}
\tagLoss{GISTEmbedLoss}\tagLLM{Pointwise reranking}
\tagLLM{Pairwise reranking}
\tagLLM{Tournament pairwise reranker} 
\tagLLM{Hypothetical Document Embeddings Query Expansion}&
\tagModel{JobBERT-v2}
\tagModel{Qwen-2.5-7B-Instruct-AWQ} \\

\midrule

baorphuc &
\cite{bao2026baorphuc} &
\tagModelType{Encoder}\tagTechniques{Retrieval}\tagLoss{InfoNCE} &
\tagModel{all-mpnet-base-v2} \\

\midrule

classum &
\cite{kim2026classum} &
\tagModelType{Encoder}\tagModelType{Decoder}\tagTechniques{Retrieval}\tagTechniques{Reranking}
\tagLoss{Batch contrastive loss}\tagLoss{CachedGISTEmbedLoss}
\tagLoss{GISTEmbedLoss}\tagLoss{Encoder Fusion}
\tagLLM{Prompt engineering}\tagLLM{Data Augmentation}
\tagExternalData{External data}&
\tagModel{ZEmbed1-4B}
\tagModel{KaLM-12B}
\tagModel{Qwen3-8B}
\tagModel{NVEmbed-3B}
\tagModel{Zerank2} \\

\midrule

MARSAD &
\cite{ibrahim2026marsad} &
\tagModelType{Encoder}\tagModelType{Lexical}\tagTechniques{Retrieval}\tagTechniques{Rank fusion}&
\tagModel{BM25}
\tagModel{intfloat/e5-base-v2} \\

\midrule

Olive &
\cite{ranjan2026olive} &
\tagModelType{Encoder}\tagModelType{Lexical}\tagTechniques{Retrieval}
\tagLoss{Curriculum Learning}\tagLoss{InfoNCE}
\tagLoss{Reciprocal Rank Fusion} &
\tagModel{BM25}
\tagModel{BAAI/bge-base-en-v1.5}
\tagModel{paraphrase-multilingual-mpnet-base-v2} \\

\midrule

Skillberg.app &
\cite{vielvoye2026skillberg} &
\tagModelType{Encoder}\tagTechniques{Retrieval}\tagTechniques{Classifier}
\tagLoss{Proprietary job-skill knoweledge graph}
\tagLoss{Cached InfoNCE} &
\tagModel{intfloat/multilingual-e5-large-instruct} \\

\midrule

ui-nlp &
\cite{luthfiyyah2026uinlp} &
\tagModelType{Encoder}\tagTechniques{Retrieval}\tagLoss{InfoNCE} &
\tagModel{intfloat/multilingual-e5-large-instruct}
 \\

\midrule

hr\_gradient &
\cite{castaneira2026semanticreranking} &
\tagModelType{Encoder}\tagTechniques{Retrieval}\tagTechniques{Reranking}
\tagLoss{Cross-encoder reranking}\tagLoss{Domain adaptation}

\tagLoss{CachedGIST}\tagLLM{Prompt Engineering}
\tagLLM{Data Augmentation}\tagExternalData{External data} &
\tagModel{Alibaba-NLP/gte-modernbert-base}
\tagModel{Qwen/Qwen3-Reranker-0.6B }
\tagModel{Gemma 4}\\

\midrule

bipboopbipboop &
\cite{hemamou2026retrievereranksurvive} &
\tagModelType{Encoder}\tagModelType{Decoder}\tagTechniques{Retrieval}\tagTechniques{Reranking}
\tagLoss{Query expansion}\tagLoss{InfoNCE}\tagLoss{LoRA}
\tagLoss{Score Fusion}\tagLLM{Listwise select-then-rank}
\tagLLM{Pointwise reranking}\tagLLM{Prompt engineering} &
\tagModel{Qwen3-Embedding-4B}
\tagModel{Qwen3-Reranker-8B}
\tagModel{Qwen3-235B-A22B-Instruct}
\tagModel{JobBERT-v3} \\

\bottomrule
\end{tabular}
}
\end{table}

As in Task A, the solutions proposed for Task B are based on information retrieval architectures that combine hybrid strategies, but with a stronger emphasis on representation learning, likely due to the large training and development sets provided. In several systems, the initial retrieval step is followed by reranking, rank fusion, or LLM-based refinement.

In the initial retrieval stage, encoder-based models are the dominant model choice. Several teams rely on general-purpose pretrained embedding models, including all-mpnet-base-v2, models from the e5 family~\cite{wang2024multilingual}, BGE models~\cite{chen2024bge}, and Qwen-based embedding models~\cite{zhang2025qwen3}. Some teams, such as \textit{NightSun} and \textit{bipboopbipboop}, also use domain-adapted models from the JobBERT family. Although lexical retrieval appears less frequently than in Task A, BM25 is still used by \textit{MARSAD} and \textit{Olive}.

A pattern in Task B is the adaptation of dense retrievers to the provided supervision data. Several teams fine-tune their models with contrastive or metric-learning objectives, with the aim of aligning the embedding space with the relevance patterns captured in the training and development sets. \textit{baorphuc}, \textit{Olive}, \textit{ui-nlp}, and \textit{bipboopbipboop} use InfoNCE losses~\cite{oord2018representation}, while \textit{Skillberg.app} applies contrastive learning with Cached InfoNCE. Similarly, \textit{classum} and \textit{NightSun} use GISTEmbedLoss~\cite{solatorio2024gistembed}. Other fine-tuning strategies include curriculum learning, used by \textit{Olive}; domain adaptation and external data, used by \textit{hr\_gradient}; and LoRA-based adaptation of large embedding models, used by \textit{bipboopbipboop}.


LLM-related methods are used in several submissions, mainly for reranking, data augmentation, and query expansion. \textit{NightSun} uses LLM-based pointwise reranking and a tournament pairwise reranker, while \textit{bipboopbipboop} combines pointwise reranking with a listwise select-then-rank strategy. In contrast, \textit{classum} and \textit{hr\_gradient} report mainly the use of LLMs for data augmentation. Generative models are also used to enrich the input before retrieval: \textit{NightSun} uses Hypothetical Document Embeddings for query expansion, and \textit{bipboopbipboop} also includes query expansion as part of their pipeline. These techniques seem especially useful when the input is short or lexically different from the target items, as is the case with the data provided.


\section{Results}
\subsection{Task A Main Results}
\subsubsection{Overall Task A Performance}

The main leaderboard, shown in Table~\ref{tab:team_results_task_a}, reports the Mean Average Precision across the monolingual English and Spanish scenarios. \textit{classum} achieved the best overall performance, with an average MAP of 0.7140, obtaining the highest results in both English--English (0.7122) and Spanish--Spanish (0.7157). \textit{QTPride} ranked second overall, with an average MAP of 0.6632, also achieving the second-best results in both languages. \textit{bipboopbipboop} completed the top three with an average MAP of 0.6532, showing balanced performance in English and Spanish. It is also notable that \textit{Skillberg.app} obtained a strong English score of 0.6608, close to the second-best result in that language, while \textit{CYUT} performed comparatively better in Spanish (0.6213) than in English (0.5974).

\begin{table}[ht!]
\centering
\caption{Overview of team results for Task A.}
\label{tab:team_results_task_a}
\resizebox{0.90\textwidth}{!}{%
\begin{tabular}{@{}l @{\hspace{0.2cm}} 
                c @{\hspace{0.2cm}}
                c @{\hspace{0.2cm}}
                c @{\hspace{0.2cm}}
                c@{}}
\toprule
\textbf{Team} & \textbf{System ID} & \textbf{Avg} & \textbf{MAP(en-en)} & \textbf{MAP(es-es)} \\
\midrule
\rowcolor{gold!20} classum              & 715941 & \textbf{0.7140} & \textbf{0.7122} & \textbf{0.7157} \\
\rowcolor{silver!20} QTPride            & 696980 & \underline{0.6632} & \underline{0.6622} & \underline{0.6641} \\
bipboopbipboop                          & 715693 & 0.6532 & 0.6552 & 0.6512 \\
Skillberg.app                             & 709705 & 0.6496 & 0.6608 & 0.6384 \\
ui-nlp                                  & 715491 & 0.6264 & 0.6327 & 0.6202 \\
CYUT                                    & 702458 & 0.6094 & 0.5974 & 0.6213 \\
DevoMatcher                             & 999999 & 0.5989 & 0.5966 & 0.6012 \\
HR\_NLP                                 & 715595 & 0.5605 & 0.5540 & 0.5669 \\
VerbaNex                                & 710446 & 0.5576 & 0.5614 & 0.5538 \\
qwerity                                 & 699840 & 0.5066 & 0.5516 & 0.4616 \\
UBCS                                    & 701328 & 0.4905 & 0.5144 & 0.4666 \\
TalentCLEF Baseline                     & 690133 & 0.3828 & 0.4001 & 0.3655 \\
TalentCLET Task1-M5-G1\footnotemark     &    -    &    -    &  -      &    -    \\
\bottomrule
\end{tabular}%
}
\end{table}
\footnotetext{This team did not have valid submissions for the test set, only for the development set.}

The two top-performing systems in Task A shared a common aspect in transforming job descriptions and resumes into more comparable semantic representations before ranking. \textit{classum} relied on the semantic enrichment of both job descriptions and resumes, including extracted skill and task profiles, which were compared with structured representations of responsibilities, seniority, domain, and role compatibility inferred from the development set. Then, they combined many dense semantic representations with LLM-based listwise scoring, rubric-based scoring, score fusion, and reranking. \textit{QTPride}, the second-ranked team, followed a related multi-view strategy: they first parsed job descriptions and resumes into structured JSON representations using an LLM, and then combined multiple retrieval views based on full-text, work-experience, and ESCO knowledge-graph through reciprocal rank fusion. A similar trend can also be observed in \textit{Skillberg.app}, which incorporated a proprietary job--skill knowledge graph as part of their training strategy. The third-ranked team, \textit{bipboopbipboop}, proposed a particularly strong multi-stage reranking pipeline: candidate resumes were first scored by complementary matchers, including neural rerankers, bi-encoders, and an LLM-as-a-judge system, then combined through geometric-mean rank fusion, and finally refined with an LLM-based listwise tournament over the top candidates. The top-performing systems suggest that job--person matching benefits from going beyond plain-text similarity: structured document understanding, extracted entities, graph-derived relations, fusion mechanisms, and selective reranking help capture different dimensions of candidate--job alignment.

\subsubsection{Cross-Lingual Performance} To evaluate performance in the cross-lingual setting, the metric considered was MAP for the English--Spanish language pair, where English queries were matched against Spanish corpus elements. The results are presented in Table~\ref{tab:team_results_task_a_cross_lingual}. \textit{classum} also achieved the best cross-lingual performance, reaching a MAP of 0.7044. \textit{bipboopbipboop} ranked second, with a MAP of 0.6438, followed closely by \textit{Skillberg.app} with a MAP of 0.6373. \textit{QTPride} also obtained a competitive score of 0.6355, indicating that the top systems performed similarly in this setting. 

The strongest cross-lingual systems followed patterns similar to those observed in the multilingual setting. The best-performing system, \textit{classum}, combined multiple cross-lingual dense representations with enriched job and resumes views. \textit{bipboopbipboop}, the second-ranked team in this benchmark, adapted their pipeline to the cross-lingual setting by using only two of their four matchers: a multilingual BGE-M3 bi-encoder and a neural reranker. In this configuration, it is likely that BGE-M3 provided an important multilingual semantic signal within the final fused representation, which was subsequently refined through an LLM-based listwise tournament. Interestingly, for the teams at the top of the benchmark, the differences between the cross-lingual and multilingual results were relatively small, suggesting that these systems were able to generalize effectively to the English--Spanish matching scenario.

\begin{table}[ht!]
\centering
\caption{Overview of team results for Task A cross-lingual setting. Best value per column is in bold, second best is underlined.}
\label{tab:team_results_task_a_cross_lingual}
\resizebox{0.60\textwidth}{!}{%
\begin{tabular}{@{}l @{\hspace{0.2cm}} 
                c @{\hspace{0.2cm}}
                c@{}}
\toprule
\textbf{Team} & \textbf{System ID} & \textbf{MAP(en-es)} \\
\midrule
\rowcolor{gold!20} classum              & 707960 & \textbf{0.7044} \\
\rowcolor{silver!20} bipboopbipboop     & 715693 & \underline{0.6438} \\
Skillberg                               & 690318 & 0.6373 \\
QTPride                                 & 696980 & 0.6355 \\
ui-nlp                                  & 715491 & 0.5995 \\
CYUT                                    & 702388 & 0.5851 \\
HR\_NLP                                 & 715595 & 0.5518 \\
hr\_gradient                            & 716035 & 0.5348 \\
DevoMatcher                             & 999999 & 0.5293 \\
VerbaNex                                & 703373 & 0.4613 \\
TalentCLEF Baseline                     & 690133 & 0.3562 \\
UBCS                                    & 701251 & 0.3562 \\
\bottomrule
\end{tabular}%
}
\end{table}

\subsubsection{Bias Evaluation}
In addition to standard retrieval metrics, we performed a bias-oriented analysis by measuring the consistency of the rankings in the gendered variants of the same records. Synthetic evaluation data was created from paired examples in which only gender-marked elements, such as names and job titles, were modified, while the remaining profile information was kept unchanged. This setup makes it possible to assess whether the systems preserve similar rankings when equivalent profiles are presented with masculine or feminine variants. Systems with smaller discrepancies between these variants can therefore be considered to be more robust to this specific type of gender perturbation.

We used Rank-Biased Overlap (RBO)~\cite{webber2010similarity} to compare the rankings obtained for the gendered variants in the English--English, English--Spanish, and Spanish--Spanish scenarios. The average values are shown in Table~\ref{tab:team_results_task_a_rbo}. \textit{classum} achieved the highest overall RBO score, with an average of 0.9904, obtaining the best results in English--English (0.9931), English--Spanish (0.9892), and Spanish--Spanish (0.9890). \textit{hr\_gradient} ranked second overall, with an average RBO of 0.9547, and achieved the second-best results in English--English (0.9664) and English--Spanish (0.9489). Meanwhile, \textit{HR\_NLP} ranked third overall, with an average RBO of 0.9521, and obtained the second-best score in Spanish--Spanish (0.9538).

\begin{table}[ht!]
\centering
\caption{Overview of team results for Task A using RBO metrics. Best value per column is in bold, second best is underlined.}
\label{tab:team_results_task_a_rbo}
\resizebox{0.95\textwidth}{!}{%
\begin{tabular}{@{}l @{\hspace{0.2cm}} 
                c @{\hspace{0.2cm}}
                c @{\hspace{0.2cm}}
                c @{\hspace{0.2cm}}
                c @{\hspace{0.2cm}}
                c@{}}
\toprule
\textbf{Team} & \textbf{System ID} & \textbf{Avg} & \textbf{RBO(en-en)} & \textbf{RBO(en-es)} & \textbf{RBO(es-es)} \\
\midrule
\rowcolor{gold!20} classum              & 707960 & \textbf{0.9904} & \textbf{0.9931} & \textbf{0.9892} & \textbf{0.9890} \\
\rowcolor{silver!20} hr\_gradient       & 716035 & \underline{0.9547} & \underline{0.9664} & \underline{0.9489} & 0.9488 \\
HR\_NLP                                 & 715595 & 0.9521 & 0.9558 & 0.9466 & \underline{0.9538} \\
Skillberg                               & 709705 & 0.9338 & 0.9369 & 0.9309 & 0.9336 \\
DevoMatcher                             & 999999 & 0.9091 & 0.9492 & 0.8667 & 0.9114 \\
QTPride                                 & 696980 & 0.8842 & 0.8971 & 0.8575 & 0.8979 \\
bipboopbipboop                          & 715693 & 0.8727 & 0.8714 & 0.8631 & 0.8837 \\
CYUT                                    & 702388 & 0.8581 & 0.8528 & 0.8421 & 0.8795 \\
UBCS                                    & 701328 & 0.8211 & 0.8624 & 0.7665 & 0.8344 \\
ui-nlp                                  & 715491 & 0.8110 & 0.8251 & 0.7789 & 0.8291 \\
TalentCLEF Baseline                     & 690133 & 0.7854 & 0.7919 & 0.7800 & 0.7843 \\
VerbaNex                                & 709530 & 0.6960 & 0.7580 & 0.5848 & 0.7453 \\
\bottomrule
\end{tabular}%
}
\end{table}

The high RBO scores obtained by \textit{classum} indicate that their rankings were highly stable under the gendered perturbations considered in this evaluation. A plausible explanation is that, by extracting task and skill-oriented profiles, the system likely reduced the influence of surface-level lexical changes in names or gender-marked job titles. In addition, their use of multiple dense embedding models, LLM-based scoring components, score fusion, and reranking may have further stabilized the final ranking by aggregating complementary signals rather than relying on a single textual representation that might be affected more by gender-bias. However, none of the top systems explicitly report a dedicated bias-mitigation component for this setting, and pretrained embedding models or LLMs may still inherit biases from their underlying data. Therefore, these results should be interpreted as evidence of ranking robustness under the specific masculine/feminine variants used in the benchmark, rather than as a comprehensive demonstration of fairness or bias mitigation.

\subsection{Task B Main Results}

The main Task B leaderboard, shown in Table~\ref{tab:team_results_task_b_ndcg}, reports the NDCG results using both graded and binary relevance. Rows are ordered by NDCG(graded), the main evaluation metric. \textit{classum} achieved the best overall performance, with an NDCG(binary) of 0.8340 and an NDCG(graded) of 0.8068, obtaining the highest results in both metrics. \textit{NightSun} ranked second overall, with an NDCG(binary) of 0.8246 and an NDCG(graded) of 0.7913, also achieving the second-best results in both metrics. \textit{bipboopbipboop} completed the top three, with a NDCG(binary) of 0.8123 and a NDCG(graded) of 0.7793.

\begin{table}[ht!]
\centering
\caption{Overview of team results for Task B using graded and binary NDCG metrics. Rows are ordered by NDCG(graded), the main evaluation metric. Best value per metric column is in bold, second best is underlined.}
\label{tab:team_results_task_b_ndcg}
\resizebox{0.80\textwidth}{!}{%
\begin{tabular}{@{}l @{\hspace{0.2cm}}
                c @{\hspace{0.2cm}}
                c @{\hspace{0.2cm}}
                c@{}}
\toprule
\textbf{Team} & \textbf{System ID} & \textbf{NDCG(graded)} & \textbf{NDCG(binary)} \\
\midrule
\rowcolor{gold!20} classum          & 715319 & \textbf{0.8068} & \textbf{0.8340} \\
\rowcolor{silver!20} NightSun       & 709648 & \underline{0.7913} & \underline{0.8246} \\
bipboopbipboop                      & 714675 & 0.7793 & 0.8123 \\
hr\_gradient                        & 716124 & 0.7399 & 0.7746 \\
Skillberg.app                       & 710631 & 0.7088 & 0.7487 \\
baorphuc                            & 692732 & 0.6829 & 0.7285 \\
ui-nlp                              & 716152 & 0.6821 & 0.7278 \\
MARSAD                              & 697840 & 0.6531 & 0.6950 \\
Olive                               & 693022 & 0.6458 & 0.6899 \\
TalentCLEF Baseline                 & 690199 & 0.6204 & 0.6634 \\
\bottomrule
\end{tabular}%
}
\end{table}

The approaches used by the best-performing systems in Task B show a clear predominance of fine-tuned dense retrieval methods combined with reranking and representation enrichment. \textit{classum}, the top-ranked system, combined several encoder models, some of them fine-tuned using contrastive learning objectives such as GISTEmbedLoss and CachedGISTEmbedLoss, and then applied a Zerank2 reranker~\cite{pipitone2025zelo}. Their system also used prompt engineering for data augmentation and generated enriched job title and skill-concept views, allowing the model to compare short job titles and ESCO skills through more informative representations. \textit{NightSun}, the second-ranked team, also followed a retrieval-and-reranking strategy, combining a fine-tuned encoder with query augmentation based on hypothetical document embeddings, followed by pointwise and pairwise reranking. \textit{bipboopbipboop}, ranked third, used a multi-stage pipeline in which job titles were expanded before retrieval, candidates were retrieved using a fine-tuned Qwen3 embedding model, the candidate pool was enriched with JobBERT-v3 as a domain-specialized model, and the final ranking was refined through pointwise and listwise LLM-based reranking. 

In a task where the available textual context was relatively limited, embedding fine-tuning and data augmentation techniques seem to have played an important role in improving system performance. In addition, the use of recent neural rerankers and LLM-based reranking strategies, beyond traditional cross-encoder reranking, may have contributed substantially to the performance of the strongest systems.

\section{Conclusions}

TalentCLEF 2026 consolidated the first community evaluation campaign focused on NLP in Human Capital Management. The second edition attracted substantial participation, with 113 registered participants, 29 teams submitting at least one run to the official benchmarks, and 17 teams contributing system papers. Across the two tasks, the challenge received more than 400 submissions, reflecting the growing interest of the research community in reproducible benchmarks for job, candidate, and skill intelligence. The reuse of previous TalentCLEF resources by several participants, including datasets and participant outputs generated during the first edition, indicates that the initiative is beginning to support research in this domain~\cite{decorte2025multilingual}. It has also helped inspire the creation of new resources aimed at advancing research in the area~\cite{de2026workrb}.

The TalentCLEF 2026 results highlight a clear methodological trend in both tasks: the strongest systems relied on hybrid and modular architectures rather than single-model retrieval. In Task A, contextualized job-person matching was most effectively addressed as a multi-stage retrieval and reranking problem, where semantic representations of job descriptions and resumes were combined with LLM-generated structured views such as extracted skills, tasks, work experience, and, in some cases, graph-derived information. In Task B, job-skill matching followed a similar retrieval-oriented architecture, but with greater emphasis on adapting the embedding space to the task. Fine-tuned encoders trained with contrastive or metric-learning objectives were often combined with query expansion, data augmentation, representation enrichment, fusion, and reranking to distinguish between core and contextual skills.

In both tasks, the best-performing approaches show the value of combining complementary rankings. Dense encoders provided robust semantic similarity, lexical or domain-specific models captured terminology and labor-market language, and graph-based resources or extracted entity views added information not directly available from surface text. Reranking also played an important role, with both LLM-based and recent neural rerankers refining the top candidates retrieved in earlier stages.

From an evaluation perspective, the systems achieved strong results in all languages and showed different degrees of robustness in bias-oriented evaluation, even though this aspect was not explicitly addressed by most of the participants. For this reason, the next edition of the challenge will place greater emphasis on promoting and extending this type of evaluation.



%
%
%
%
\bibliographystyle{splncs04}  
\bibliography{bibliography}

\end{document}